\documentclass[conference]{IEEEtran}
\IEEEoverridecommandlockouts
\usepackage{cite}
\usepackage{amsmath,amssymb,amsfonts}
\usepackage{algorithmic}
\usepackage{algorithm}
\usepackage{graphicx}
\usepackage{textcomp}
\usepackage{xcolor}
\usepackage{enumitem}
\def\BibTeX{{\rm B\kern-.05em{\sc i\kern-.025em b}\kern-.08em
    T\kern-.1667em\lower.7ex\hbox{E}\kern-.125emX}}
\begin{document}

\title{Agentic Router: An Execution-Grounded Continual Learning Approach With Memory}

\author{Yuxuan Chen, Rongpeng Li, Zhifeng Zhao, Yuntao Liu, Xing Xu and Honggang Zhang
\thanks{Y. Chen and R. Li are with Zhejiang University, Hangzhou 310027, China (email: \{cyx00, lirongpeng\}@zju.edu.cn). Z. Zhao is with Zhejiang Lab, Hangzhou 310012, China, as well as Zhejiang University, Hangzhou 310027, China (email: zhaozf@zhejianglab.org). Y. Liu is with Zhejiang Lab, Hangzhou 310012, China (email: liuyt@zhejianglab.org). X. Xu is with State Grid Hebei Electric Power Co., Ltd., Hebei, China (email: 836240885@qq.com). H. Zhang is with Macau University of Science and Technology, Macau, China (email: hgzhang@must.edu.mo).}}
\maketitle

\begin{abstract}
Large language model (LLM) agents provide a promising interface for command-line-based network operations, but a plausible command may still fail or introduce operational risk after execution. Existing approaches mainly focus on command generation or final configuration correctness, and do not use execution-grounded experience to jointly improve candidate coverage and action selection. We propose an execution-grounded dual-path consequence-aware agent for CLI-based SONiC operations, which generates multiple complete actions, predicts their execution consequences, and selects the final action through utility- and risk-aware reranking. The proposal-side path abstracts reusable operational lessons into retrievable guidance to improve feasible-action coverage without modifying the proposal LLM, while the selection-side path adapts the consequence predictor through session-level LoRA updates using real SSH feedback to improve conditional selection quality. Experiments over multi-turn SONiC operation sessions with different Qwen3 proposal models show that the framework improves feasible-action coverage and top-1 execution success, and that the two adaptation paths provide complementary gains over interaction.
\end{abstract}


\section{Introduction}
Modern networks increasingly rely on open and programmable systems such as SONiC~\cite{sonicproject2026}, where many operational tasks are performed through command-line interfaces (CLIs). 
Recent progress in large language model (LLM) agents~\cite{yao2023react} brings a new opportunity for this process. An operator can describe an intent in natural language, and an agent can propose executable network actions. This idea is also aligned with the broader trend of intent-based networking~\cite{leivadeas2023survey} and LLM-assisted network management~\cite{liu2025llmnetworksurvey,NetGPT}.
However, CLI operation is not a static text-generation problem. A command may be syntactically valid and executable yet remain inappropriate for the current network state. For example, when only one Border Gateway Protocol (BGP) peer requires recovery, the CLI \texttt{clear bgp *}, which clears all peers, may cause route churn and transient reachability loss. 
Such a failure cannot be identified from linguistic plausibility alone; it becomes observable only through command responses and post-execution state verification.
Reliable CLI agents therefore need to consider not only whether plausible actions can be generated, but also whether the action selected for execution has a satisfactory consequence under the observed device state.

Recent LLM-based network automation studies~\cite{NetConfEval,wan2025netkeeper} have advanced intent-to-configuration translation and automated network updates. However, their LLM components mainly generate or validate outputs for the current task, rather than accumulate verified device-side outcomes to improve subsequent operations. Beyond networking, interactive and embodied agents formulate decision-making as an action--observation loop. They select actions according to both task relevance and environmental feasibility, and use execution feedback to guide later decisions or retain reusable experience~\cite{yao2023react,ichter2023saycan,shinn2023reflexion}. 
In interactive settings, agents should select complete actions according to their expected outcomes in the environment, rather than linguistic plausibility alone. These outcomes indicate whether an action can be executed to advance the task. 
However, existing methods do not provide an execution-grounded mechanism that uses such verified CLI outcomes to support reliable and continual improvement over multi-turn interaction in a stateful, environment-dependent network system.

CLI-based network operations with feedback can be modeled as a multi-turn action-level decision process, where observed consequences provide evidence for later adaptation. However, a failed top-ranked execution is ambiguous for adaptation: it does not reveal whether the candidate set lacks a feasible action or whether an available feasible action is not ranked first. This ambiguity motivates separating candidate action proposal from consequence-aware action selection, since the accompanied \emph{top-1 failure} corresponds to different adaptation targets. Proposal-side omissions often reflect missing reusable knowledge about command patterns, CLI constraints, and verification procedures. 
Such knowledge should be explicit and reusable across tasks, rather than retained only in transient hidden states or raw histories~\cite{dai2019transformerxl}. Retrieval-based memory ~\cite{lewis2020rag,ouyang2026reasoningbank} is therefore suitable for preserving execution-validated operational lessons and injecting them into future proposals.
But, retrieval does not update the predictor's state--action consequence mapping. It therefore cannot directly correct session-specific misranking errors among available candidates.
Full-model online updating~\cite{online_continual_learning,test_time_training} can be useful, but network-operation feedback is state-dependent and safety-critical. An update learned from one CLI context may disturb previously reliable judgments about other valid command patterns or risk levels. Low-rank adaptation (LoRA) instead provides session-level plasticity while keeping the base predictor stable~\cite{hu2022lora}. 

Based on these observations, we propose an execution-grounded dual-path adaptation framework for CLI-based network operations. The framework separates the candidate proposal from consequence-aware action selection. A proposal LLM first generates multiple complete candidate actions. A consequence predictor estimates their likely execution consequences, and utility- and risk-aware reranking selects the action to be executed. Observed consequences, then update the two adaptation paths. The proposal-side path abstracts reusable operational lessons into long-term memory and retrieves them as guidance for future proposals without modifying the proposal LLM. The selection-side path uses recent state--action--consequence records to apply session-level LoRA updates to the consequence predictor, correcting later reranking decisions. Experiments in multi-turn SONiC operation sessions show that the framework improves feasible-action coverage and top-1 execution success, and that the two adaptation paths provide complementary gains over interaction.

This work is structured as follows. Section~II presents the system model and formulates the consequence-aware decision problem. Section III details the proposed dual-path agent design and its proposal-side and selection-side adaptation mechanisms. Section~IV evaluates the framework over multi-turn SONiC operation sessions, and Section~V concludes this paper.



\section{System Model and Problem Formulation}
\label{sec:problem}

\subsection{System Model}

We consider a multi-turn network operation session in a SONiC environment. Without loss of generality, at turn $t$, the agent receives a natural-language user intent $q_t$ and an environment observation $o_t$. For example, $q_t$ can be \texttt{check counters on Ethernet76}, while $o_t$ may record the CLI context, recent command feedback, and the \texttt{EthernetN} interface naming rule. The observation $o_t$ summarizes the current device and interaction state available before action selection. Let $H_{t-1}$ denote the interaction history before turn $t$, and the information available for the current decision can be summarized as
\begin{equation}
x_t = (q_t,o_t,H_{t-1}).
\label{eq:decision_context}
\end{equation}

A basic intent-to-action agent maps the decision context to a single complete CLI action $a_t$:
\begin{equation}
a_t = G_{\psi}(x_t),
\label{eq:single_action}
\end{equation}
where $G_{\psi}$ denotes the action generator, while the action $a_t$ may consist of a single command or a short executable command sequence. After executing $a_t$, the underlying SONiC environment $\mathcal{E}_t$ returns its actual consequence $z_t$, and a task-specific verifier $y_t$ determines whether the operation succeeds:
\begin{equation}
z_t = \operatorname{Exec}_{\mathcal{E}_t}(a_t),
\qquad
y_t = v(x_t,z_t) \in \{0,1\}.
\label{eq:execution_outcome}
\end{equation}
The consequence $z_t$ records the device-side response produced by executing $a_t$, including command outputs, error messages, or observed state changes. 
For the counter-query example, the action \texttt{show interfaces counter Ethernet76} yields an observed consequence containing the device error \texttt{Unknown command}; the verifier therefore assigns $y_t=0$. By contrast, \texttt{show interfaces counters -i Ethernet76} yields the expected interface-counter output, giving $y_t=1$.

Thus, $(x_t,a_t,z_t,y_t)$ forms a basic execution record. It provides environment-grounded evidence about the state-dependent consequence of an action. However, a static single-action mapping does not indicate whether a failure is caused by not proposing a feasible action or by selecting an unsuitable one. This motivates a multi-candidate formulation that separates candidate availability from action selection.

\begin{figure*}[t]
	\centering
	\includegraphics[width=0.98\textwidth]{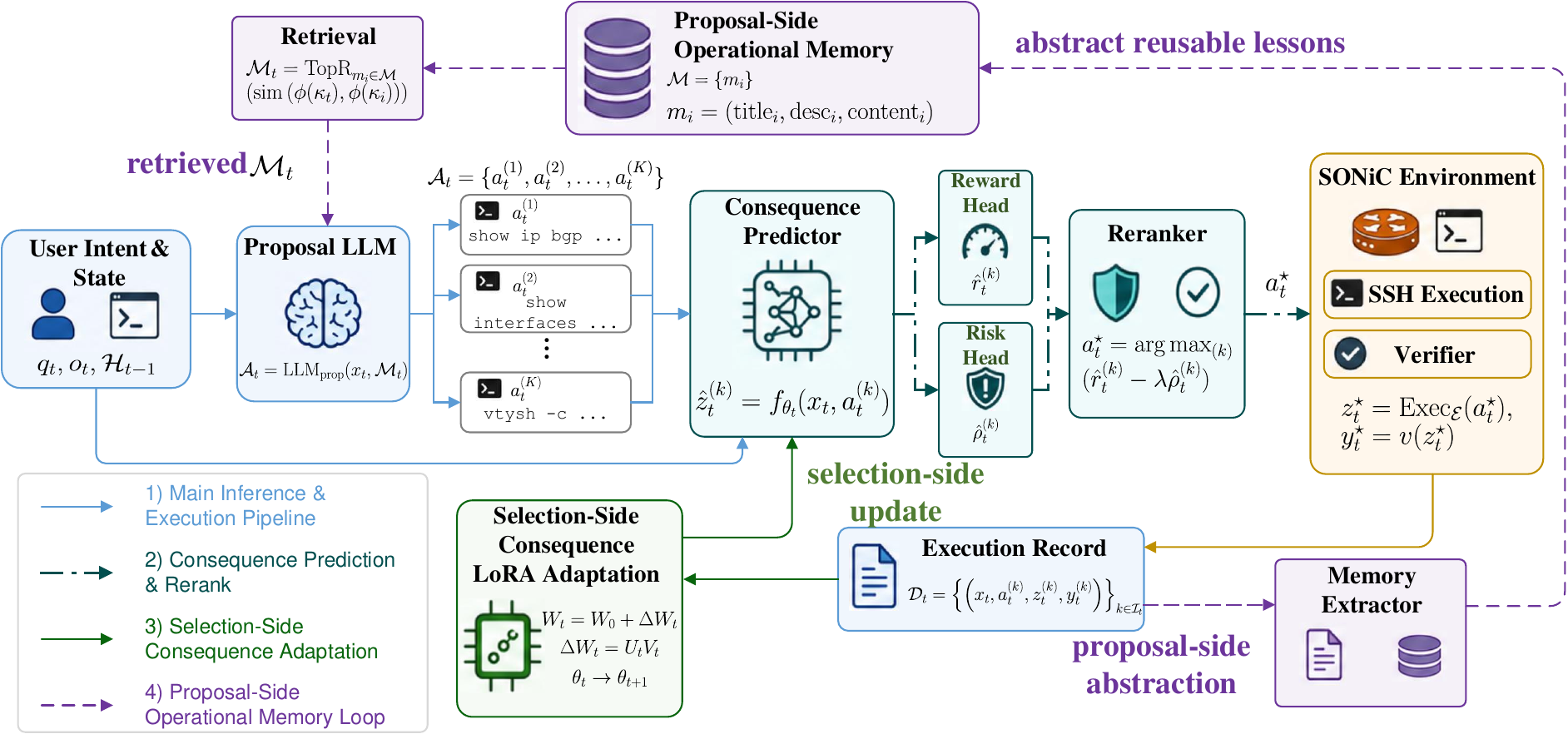}
	\caption{Overall framework of the proposed consequence-aware agent with proposal-side and selection-side adaptation for CLI-based SONiC operations.}
	\label{fig:framework}
\end{figure*}

\subsection{Problem Formulation}
\label{subsec:multi_model}
To reduce dependence on a single generation, the agent considers a set of complete candidate actions at each turn. Let $\mathcal{M}_t$ denote the proposal-side operational guidance available for the current decision, which will be discussed lately. Given the decision context $x_t$ and $\mathcal{M}_t$, the proposal generator produces
\begin{equation}
\mathcal{A}_t
=
G_{\mathrm{prop}}(x_t,\mathcal{M}_t)
=
\left\{
a_t^{(1)},a_t^{(2)},\ldots,a_t^{(K)}
\right\}.
\label{eq:candidate_generation}
\end{equation}
The candidate set expands the available action space, but it does not determine which action should be executed. 
Since the suitability of a CLI action depends on its device-side response, the agent estimates the consequence of each context--action pair:
\begin{equation}
\hat{z}_t^{(k)}
=
f_{\theta_t}
\left(
x_t,a_t^{(k)}
\right),
\qquad
a_t^{(k)}\in\mathcal{A}_t,
\label{eq:predicted_consequence}
\end{equation}
where $\hat{z}_t^{(k)}$ is the pre-execution estimate of the consequence that would be produced by executing $a_t^{(k)}$ under context $x_t$, and $\theta_t$ denotes the predictor parameters at turn $t$. These estimates provide the basis for scoring and reranking candidates before the selected action is submitted to the environment. 
Two scoring functions map the predicted consequence to expected task progress and operational risk:
\begin{equation}
\hat{r}_t^{(k)}
=
h_r\left(\hat{z}_t^{(k)}\right),
\qquad
\hat{\rho}_t^{(k)}
=
h_{\rho}\left(\hat{z}_t^{(k)}\right).
\label{eq:utility_risk}
\end{equation}
Here, operational risk refers to irreversible or highly disruptive changes to the network state. For example, when only one BGP peer requires recovery, \texttt{clear bgp *} should be assigned a high risk score because it may reset unrelated peers and cause transient route churn.
The top-ranked action under the predicted score is selected as
\begin{equation}
a_t^{\star}
=
\arg\max_{a_t^{(k)}\in\mathcal{A}_t}
\left(
\hat{r}_t^{(k)}
-
\lambda\hat{\rho}_t^{(k)}
\right),
\label{eq:top1_selection}
\end{equation}
where $\lambda\geq 0$ controls the risk penalty. 
After selection, only the top-ranked action $a_t^{\star}$ is submitted for primary execution and evaluation:
\begin{equation}
z_t^{\star}
=
\operatorname{Exec}_{E_t}
\left(a_t^{\star}\right),
\qquad
y_t^{\star}
=
v\left(x_t,z_t^{\star}\right).
\label{eq:top1_outcome}
\end{equation}
Accordingly, $z_t^\star$ and $y_t^\star$ are its observed consequence and verified outcome. This equation defines the per-turn top-1 outcome used to evaluate whether the agent's selected action succeeds.

The operational objective is to improve verified top-1 success over a session of $T$ interaction turns:
\begin{equation}
\max_{\pi}\ J_T(\pi)
=
\frac{1}{T}
\sum_{t=1}^{T}
\mathbb{E}_{\pi}
\left[
y_t^{\star}
\right],
\label{eq:session_objective}
\end{equation}
where $\pi$ denotes the overall decision and adaptation policy. 
Maximizing $J_T(\pi)$ requires reducing top-1 failures, i.e., events with $y_t^{\star}=0$. Such failures can arise at two distinct stages. Consider a request to inspect the detailed BGP neighbor state of \texttt{10.0.0.1}. If $\mathcal{A}_t$ contains only non-completing candidates, such as the unsupported form \texttt{show bgp neighbor 10.0.0.1} or the coarse summary command \texttt{show ip bgp summary}, while omitting \texttt{show ip bgp neighbors 10.0.0.1}, then no reranking rule can select a task-completing action. This is a proposal-side coverage failure. In contrast, if the detailed neighbor query is included in $\mathcal{A}_t$ but the reranker places \texttt{show ip bgp summary} above it, the selected command may execute successfully yet fail to answer the requested neighbor-level question. This is a selection-side misranking failure.

To separate these two failure sources, we define the feasible-action coverage indicator as
\begin{equation}
c_t
=
\mathbb{I}
\left[
\exists\,a_t^{(k)}\in\mathcal{A}_t
\ \text{s.t.}\
v\left(x_t,
\operatorname{Exec}_{E_t}
\left(a_t^{(k)}\right)
\right)=1
\right].
\label{eq:coverage_indicator}
\end{equation}
Thus, $c_t=1$ when the candidate set contains at least one action that can complete the current operation step. The indicator is used to diagnose candidate-set feasibility and to separate proposal coverage from the final top-ranked outcome.

Since a successful top-ranked action implies feasible-action coverage,
\begin{equation}
\Pr\left(y_t^{\star}=1\right)
=
\Pr\left(c_t=1\right)
\Pr\left(
y_t^{\star}=1
\,\middle|\,
c_t=1
\right).
\label{eq:success_decomposition}
\end{equation}
Eq. \eqref{eq:success_decomposition} implies two adaptation targets. On the proposal side, the agent should adapt the operational guidance $\mathcal{M}_t$ supplied to $G_{\mathrm{prop}}$, so that the generated candidate set is more likely to contain a feasible action, improving $\Pr\left(c_t=1\right)$. On the selection side, the agent should adapt the consequence predictor parameters $\theta_t$, so that the predicted consequences and derived scores rank an available feasible action higher (a larger $\Pr\left(y_t^{\star}=1
\,\middle|\,c_t=1\right)$). We will discuss how to implement execution-grounded dual-path adaptation in the next section.

\begin{figure*}[t]
	\centering
	\includegraphics[width=0.975\textwidth]{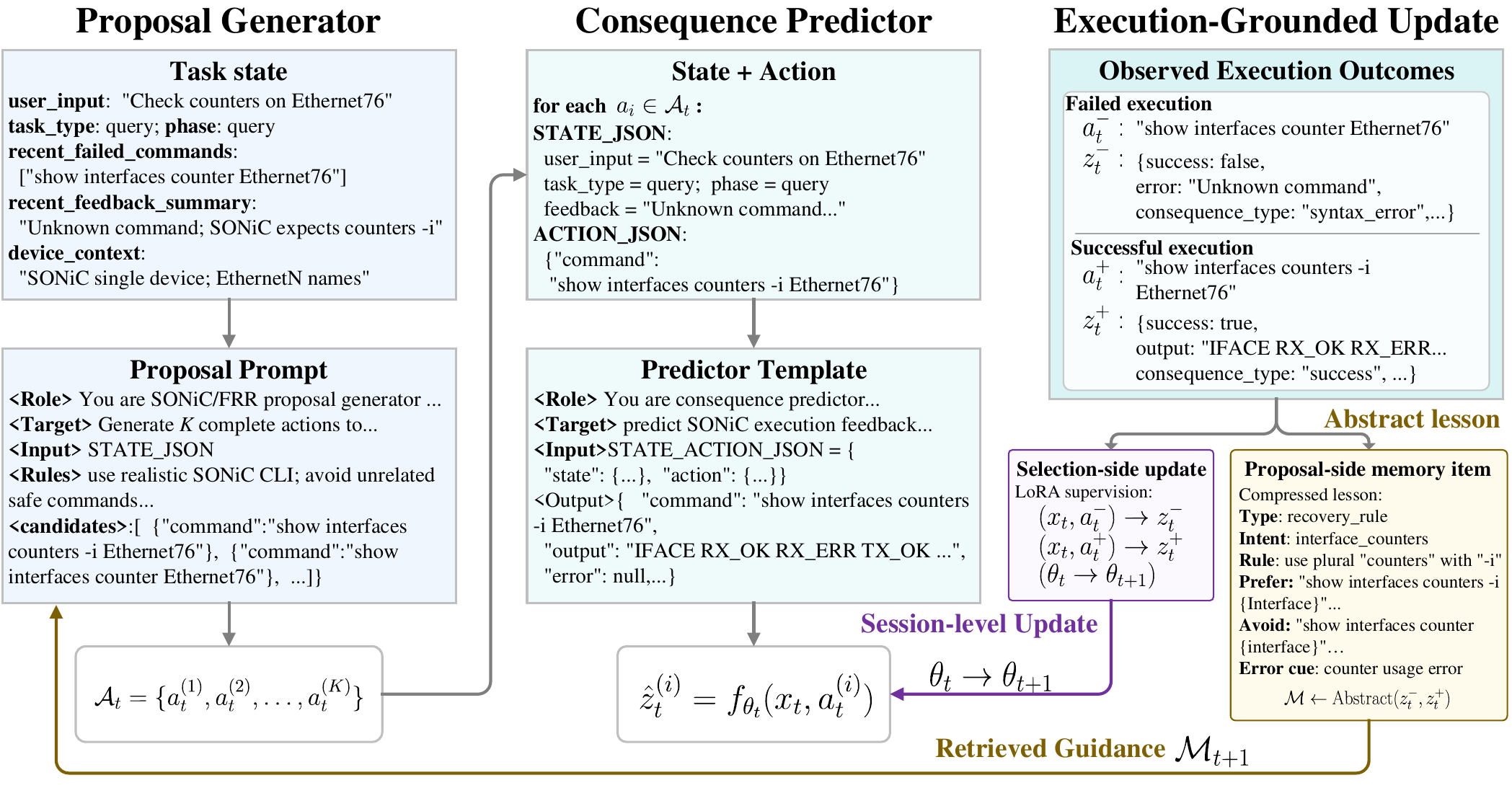}
        \caption{Prompt formulations and execution-grounded evidence routing for proposal-side memory and selection-side predictor adaptation.}
        \label{fig:prompt_template}
\end{figure*}

\section{Agentic Design with Execution-Grounded Continual Learning}
\label{sec:method}

Building on the coverage--selection decomposition in Section \ref{subsec:multi_model}, we instantiate the two adaptation targets through an execution-grounded dual-path framework. 

\subsection{Operation Loop and Execution Record Collection}
\label{subsec:pipeline}

As stated in Section \ref{subsec:multi_model}, at turn $t$, the agent first retrieves a small set of proposal-side operational guidance $\mathcal{M}_t$ according to the current decision context $x_t$. The proposal prompt then asks the LLM to return $K$ complete and distinct CLI-level actions in a structured candidate list $\mathcal{A}_t$. 
The generated candidates are then evaluated by the consequence predictor and reranked according to Eq.~\eqref{eq:top1_selection}.
Afterward, some selected action $a_t^{\star}$ is executed in the SONiC environment through SSH, and the verifier returns its actual consequence $z_t^\star$ and outcome $y_t^\star$. If the top-ranked action fails, the agent may execute lower-ranked candidates to collect additional execution feedback. 
Let $\mathcal{I}_t$ denote the indices of all candidates actually executed at turn $t$. As illustrated in Fig.~\ref{fig:prompt_template}, $\mathcal{I}_t$ consists of a failed command \texttt{show interfaces counter Ethernet76} and a successful command \texttt{show interfaces counters -i Ethernet76}, further constituting a set of execution records
\begin{equation}
\mathcal{D}_t
=
\left\{
\left(
x_t,
a_t^{(k)},
z_t^{(k)},
y_t^{(k)}
\right)
\right\}_{k\in\mathcal{I}_t}.
\label{eq:execution_record}
\end{equation}
Notably, these additional executions are used only for adaptation and do not change the top-1 outcome $y_t^{\star}$. Afterward, $\mathcal{D}_t$ is appended to the memory update buffer $\mathcal{B}$. 

\subsection{Execution-Grounded Dual-Path Adaptation}
As explained in Eq.~\eqref{eq:success_decomposition}, the collected execution records $\mathcal{D}_t$ will contribute to the dual-path adaptation: the proposal-side path updates reusable operational guidance $\mathcal{M}_t$ for candidate generation in Eq. \eqref{eq:candidate_generation}, while the selection-side path updates the consequence predictor (i.e., $h_r$ and $h_\rho$) used for reranking in Eq. \eqref{eq:utility_risk}.

\subsubsection{Proposal-Side Operational Memory}
To ensure essential guidance in limited-length prompts, raw execution feedback shall be distilled into a concise description that preserves transferable command patterns and failure conditions. Following reasoning-memory abstraction~\cite{ouyang2026reasoningbank}, a lightweight LLM-based memory extractor performs this distillation and summarizes the buffered successful, failed, and recovery operations into structured operational lessons. 
For example, from the failed command \texttt{show interfaces counter Ethernet76} and the successful recovery command \texttt{show interfaces counters -i Ethernet76}, the extractor abstracts a reusable lesson: for interface-counter queries on \texttt{EthernetN} interfaces, use the plural \texttt{counters} command with the \texttt{-i \{Interface\}} option, and avoid singular or direct-suffix variants such as \texttt{show interfaces counter \{Interface\}}.
Consequently, the proposal-side operational memory bank $\mathcal{M}$ stores compact operational lessons extracted from accumulated execution records, wherein each item $m_i\in\mathcal{M}$ contains a type, an intent description, an applicability condition, and reusable proposal guidance.

For online retrieval, the current context $x_t$ is encoded as a compact retrieval query $\kappa_t$ by the same extractor. The turn-specific guidance $\mathcal{M}_t\subseteq\mathcal{M}$ is then retrieved from the memory bank, where each memory item $m_i$ is indexed by a key $\kappa_i$ constructed from its type, intent, and applicability fields. The items are retrieved by
\begin{equation}
\mathcal{M}_t
=
\operatorname{TopR}_{m_i\in\mathcal{M}}
\operatorname{sim}
\left(
\phi(\kappa_t),
\phi(\kappa_i)
\right),
\label{eq:memory_retrieval}
\end{equation}
where $\phi(\cdot)$ is the text embedding function~\cite{sentence_bert}, $\operatorname{sim}(\cdot,\cdot)$ denotes cosine similarity, and $R$ is the number of retrieved items. The retrieved items are inserted into the proposal prompt to guide the generation of $\mathcal{A}_t$ in Eq. \eqref{eq:candidate_generation}. 

\subsubsection{Selection-Side Consequence Adaptation}
Selection-side adaptation 
updates the consequence predictor $\theta_t$ so that the reranker can better distinguish feasible actions from failed alternatives when both are available in the candidate set.

Based on state--action--consequence samples in $\mathcal{D}_t$, we implement this path as session-level LoRA adaptation~\cite{hu2022lora} of the consequence predictor. For each adapted linear layer in  $\theta_t$, we insert a low-rank update:
\begin{equation}
W_t
=
W_0+\Delta W_t,
\qquad
\Delta W_t
=
U_tV_t,
\label{eq:lora_update}
\end{equation}
where $W_0$ is the frozen base weight and $U_t,V_t$ are trainable session-level parameters. While freezing the proposal LLM and the reward and risk heads, we update $\theta_t$ by 
\begin{equation}
\mathcal{L}_{\mathrm{sel}}
=
\sum_{(x,a,z,y)\in\mathcal{D}_t}
\ell_z
\left(
f_{\theta_t}
\left(
x,a
\right),
z
\right),
\label{eq:sel_loss}
\end{equation}
where $\ell_z(\cdot,\cdot)$ measures the discrepancy between the predicted and observed consequence representations. 

Algorithm~\ref{alg:dual_memory} summarizes the resulting operation loop. Notably, the proposal-side path periodically updates $\mathcal{M}$ from accumulated records, whereas the selection-side path updates $\theta_t$ after each turn. 

\begin{algorithm}[!htbp]
\caption{Execution-grounded continual learning approach.}
\label{alg:dual_memory}
\begin{algorithmic}[1]
\REQUIRE User intent $q_t$, observation $o_t$, session history $H_{t-1}$, predictor parameters $\theta_t$, operational memory bank $\mathcal{M}$, memory update buffer $\mathcal{B}$
\STATE Construct the decision context $x_t \leftarrow (q_t,o_t,H_{t-1})$.
\STATE Retrieve $\mathcal{M}_t$ by Eq.~\eqref{eq:memory_retrieval} and generate $\mathcal{A}_t$ by Eq.~\eqref{eq:candidate_generation}.
\STATE Compute $\hat{z}_t^{(k)}$, $\hat{r}_t^{(k)}$, and $\hat{\rho}_t^{(k)}$ for all $a_t^{(k)}\in\mathcal{A}_t$.
\STATE Rank $\mathcal{A}_t$ by Eq.~\eqref{eq:top1_selection} and set $a_t^\star$ as the top-ranked action, and initialize $\mathcal{D}_t\leftarrow\emptyset$.
\STATE Execute $a_t^\star$ and evaluate $(z_t^\star,y_t^\star)$ by Eq.~\eqref{eq:top1_outcome}.
\STATE Add $(x_t,a_t^\star,z_t^\star,y_t^\star)$ to $\mathcal{D}_t$ as in Eq.~\eqref{eq:execution_record}.
\IF{$y_t^\star=0$}
    \FOR{each remaining candidate $a_t^{(k)}$ in ranked order}
        \STATE Execute $a_t^{(k)}$ by Eq.~\eqref{eq:execution_outcome}, obtaining $(z_t^{(k)},y_t^{(k)})$.
        \STATE Add $(x_t,a_t^{(k)},z_t^{(k)},y_t^{(k)})$ to $\mathcal{D}_t$ as in Eq.~\eqref{eq:execution_record}.
        \IF{$y_t^{(k)}=1$}
            \STATE \textbf{break}
        \ENDIF
    \ENDFOR
\ENDIF
\STATE Update the selection-side LoRA parameters using $\mathcal{D}_t$ and Eq.~\eqref{eq:sel_loss}, yielding $\theta_{t+1}$.
\STATE $\mathcal{B}\leftarrow\mathcal{B}\cup\mathcal{D}_t$.
\IF{the proposal-side memory update interval is reached}
    \STATE Extract reusable operational items from $\mathcal{B}$ and update $\mathcal{M}$.
    \STATE $\mathcal{B}\leftarrow\emptyset$.
\ENDIF
\STATE $H_t\leftarrow H_{t-1}\cup\{(q_t,a_t^\star,z_t^\star)\}$.
\STATE \textbf{return} Top-1 outcome $y_t^\star$, updated history $H_t$, predictor parameters $\theta_{t+1}$, and memory bank $\mathcal{M}$.
\end{algorithmic}
\end{algorithm}

\section{Experimental Results and Analysis}
\label{sec:experiments}
\subsection{Experimental Settings}
We evaluate the proposed framework over a fixed sequence of 140 interaction turns covering CLI query, configuration, and diagnosis operations. All compared methods use the same request sequence, initial environment states, candidate budget, SSH execution interface, and verifier rules. 
The agent generates $K=5$ candidate actions and evaluates the top-ranked action. We use Qwen3-1.7B, Qwen3-8B, and Qwen3-14B as proposal generators in Eq. \eqref{eq:candidate_generation}, while the consequence predictor is built on Qwen3-0.6B. 
The consequence-aware scoring module is initialized from $300$ execution records with manually annotated utility and risk labels. 
We use top-1 success as the primary metric, defined by the verified execution outcome $y_t^\star$ in Eq.~\eqref{eq:top1_outcome}. We also report Recall@$5$ as an initial proposal-side diagnostic: it measures whether the five candidates produced by the proposal generator before dual-path adaptation contain at least one feasible action, corresponding to the coverage indicator $c_t$ in Eq.~\eqref{eq:coverage_indicator}. 

We consider the following baselines as a comparison. The single-proposal baseline directly executes the first generated candidate. The self-consistency~\cite{wang2023selfconsistency} baseline samples $K=5$ candidates and selects the most frequently generated action by voting, without access to execution feedback. The no-adaptation baseline disables both proposal-side retrieval and selection-side predictor adaptation. The interaction curves report the sliding-window top-1 success rate.
\subsection{Result Analyses}

\begin{figure}[t]
    \centering
    \includegraphics[width=0.495\textwidth]{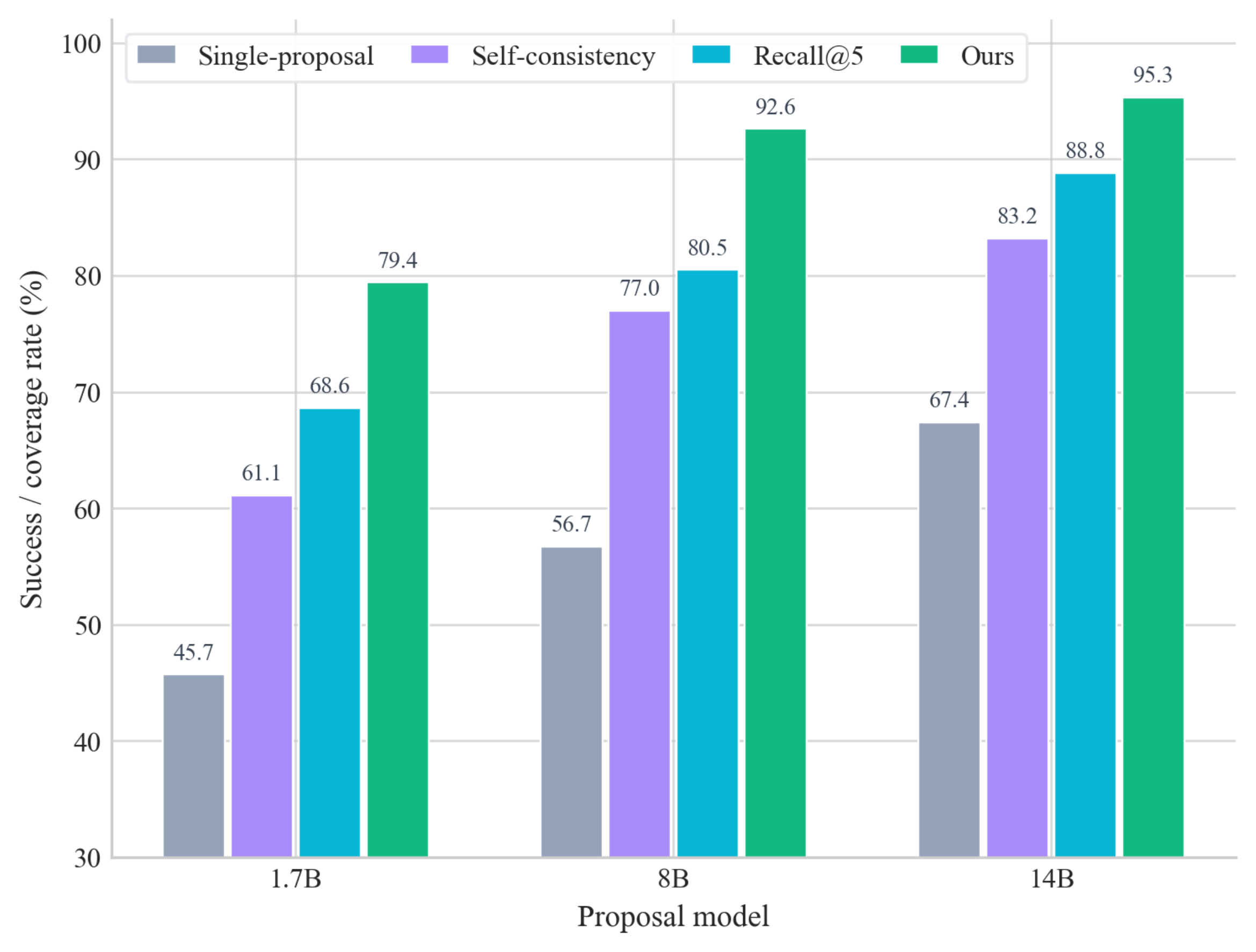}
    \vspace{-.3cm}
    \caption{Comparison of candidate-set coverage and executed top-1 success across proposal model scales.}
    \label{fig:coverage_rerank}
\end{figure}

Fig.~\ref{fig:coverage_rerank} compares the proposed method and other baselines. 
The comparison shows that a multi-candidate proposal often makes a feasible action available, and that the full dual-path framework achieves the highest top-1 success across all tested model scales. This result shows that candidate coverage alone is insufficient; consequence-aware selection and execution-grounded adaptation are still needed to convert available feasible actions into successful top-ranked executions.
\begin{figure}[t]
    \centering
    \includegraphics[width=0.495\textwidth]{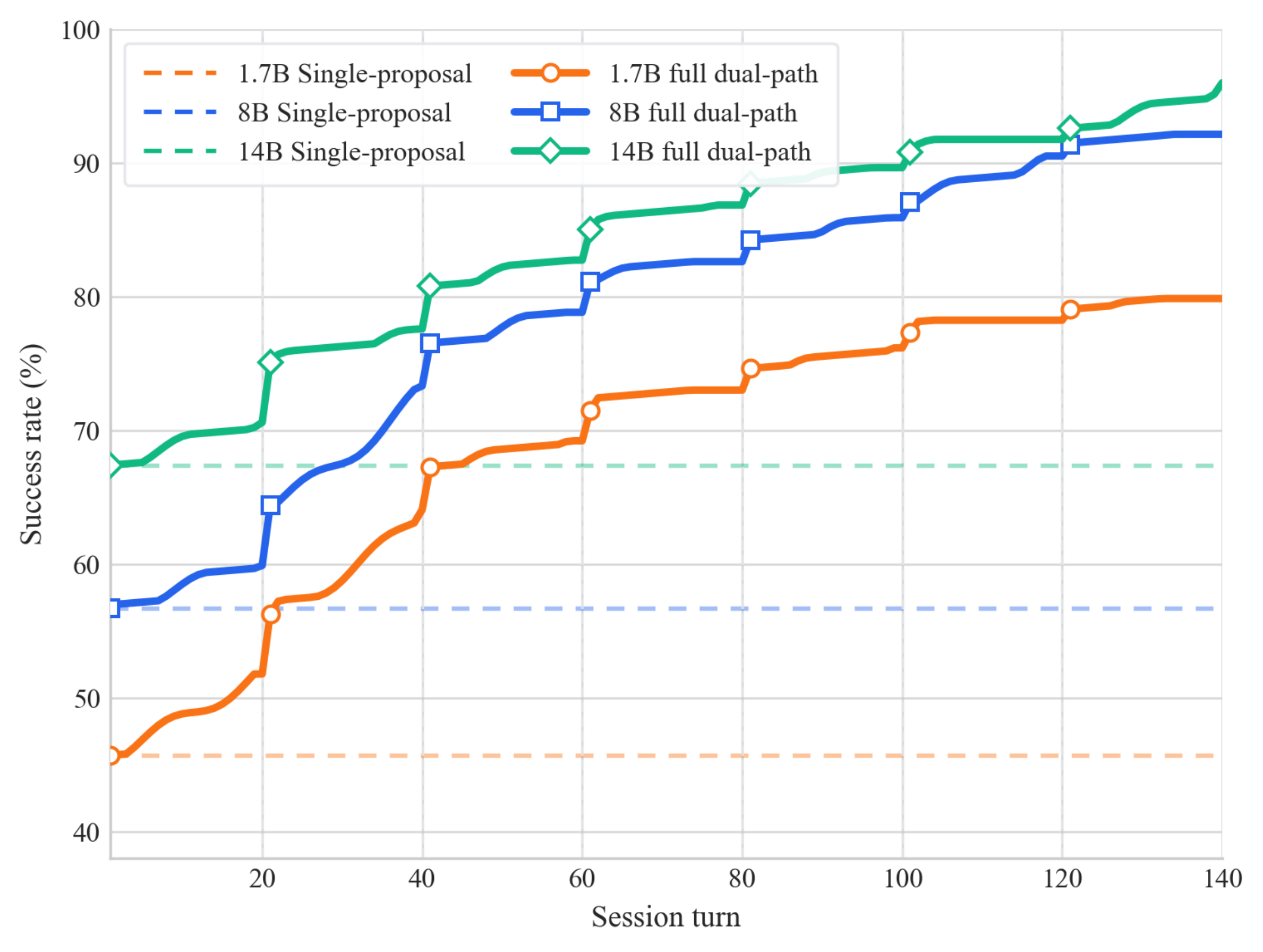}
    \vspace{-.3cm}
    \caption{End-to-end performance of the dual-path framework across proposal model scales; dashed lines denote the average no-adaptation baselines.}
    \label{fig:qwen_scale_curve}
\end{figure}

\begin{figure}[t]
    \centering
    \includegraphics[width=0.495\textwidth]{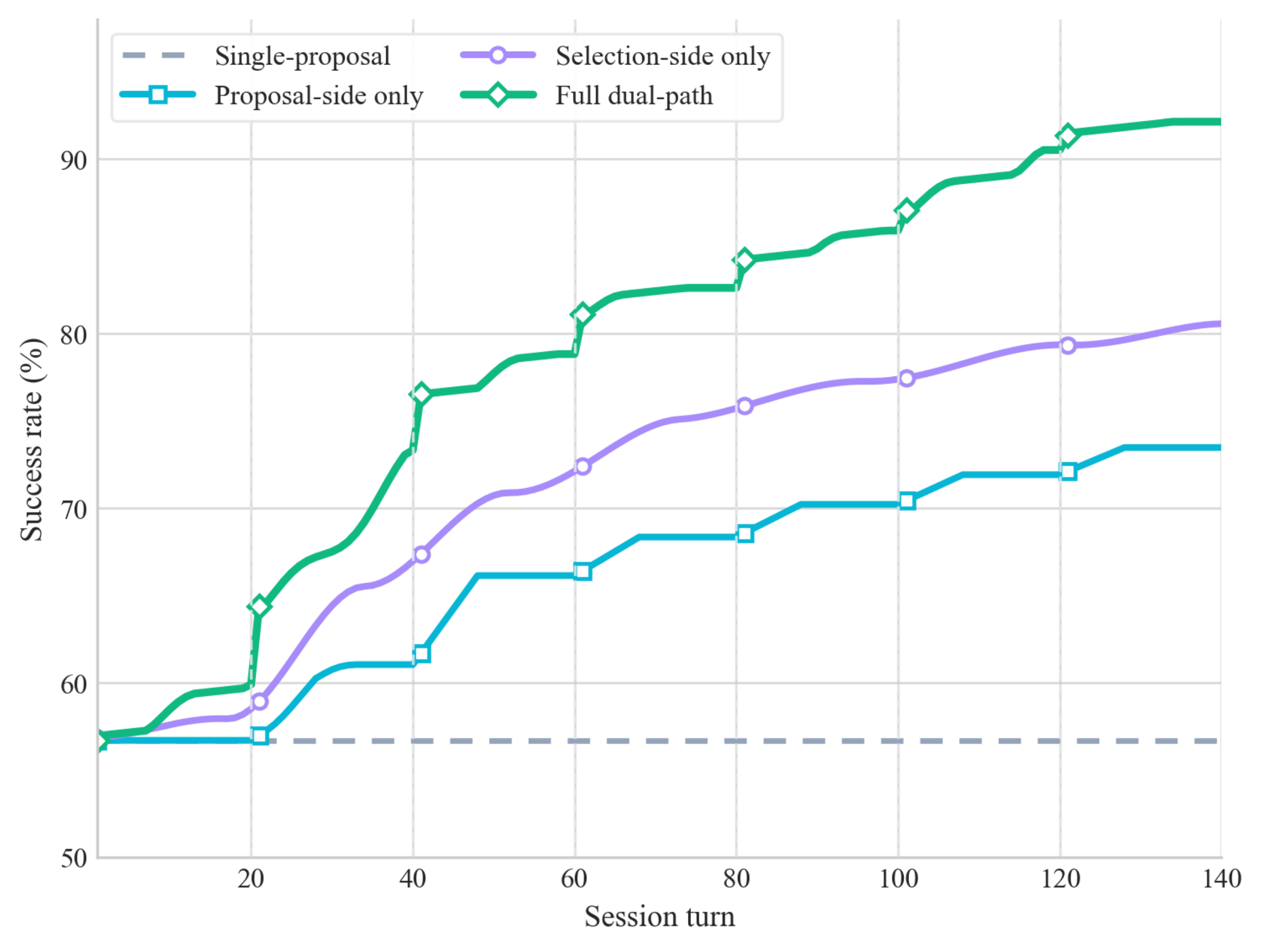}
    \caption{Adaptation-path ablation over multi-turn SONiC operation sessions.}
    \label{fig:memory_ablation}
\end{figure}

Fig.~\ref{fig:qwen_scale_curve} reports the end-to-end top-1 success of the no-adaptation and full dual-path variants over $140$ interaction turns. The full framework shows sustained improvement across Qwen3-1.7B, Qwen3-8B, and Qwen3-14B, while the larger model achieves a higher absolute performance.

Fig.~\ref{fig:memory_ablation} isolates the two adaptation paths using Qwen3-8B. The proposal-side-only variant changes only the retrieved guidance for candidate generation, and its gain is consistent with improved feasible-action coverage. The selection-side-only variant keeps proposal generation fixed and improves conditional selection quality by adapting the consequence predictor, but its performance remains bounded by the available candidates. The full framework achieves the highest success rate, supporting the decomposition in Eq.~\eqref{eq:success_decomposition}: the two adaptation paths improve complementary factors of reliable top-1 execution.

\section{Conclusion}

In this paper, we have presented an execution-grounded continual learning agent for CLI-based SONiC network operations. Guided by the feasible-action coverage and conditional selection quality formulation, the framework has adopted a dual-path adaptation. The proposal-side path has abstracted reusable operational lessons into retrievable guidance to improve feasible-action coverage without modifying the proposal LLM, while the selection-side path has adapted the consequence predictor through session-level LoRA updates using real SSH feedback to improve conditional selection quality. Experiments have shown that the proposed design has improved feasible-action coverage and top-1 execution success over multi-turn interaction, with complementary gains from the two adaptation paths. Future work will extend the framework to larger network scenarios and more comprehensive safety evaluation.



\end{document}